\documentclass{article}
\usepackage{ijcai25}

\usepackage{times}
\usepackage{soul}
\usepackage{url}
\usepackage[hidelinks]{hyperref}
\usepackage[utf8]{inputenc}
\usepackage[small]{caption}
\usepackage{graphicx}
\usepackage{amsmath}
\usepackage{amsthm}
\usepackage{booktabs}
\usepackage{algorithm}
\usepackage{algorithmic}
\usepackage[switch]{lineno}
\usepackage{array}
\usepackage{multirow}
\usepackage{colortbl}

\title{ViFusionTST: Deep Fusion of Time-Series Image Representations from Load Signals for Early Bed-Exit Prediction 
\thanks{Accepted for presentation at the AI4TS Workshop of IJCAI 2025. This work was supported by the Natural Sciences and Engineering Research Council of Canada (NSERC) under ALLRP 571106-21. (Corresponding author: Zheng Liu)}}

\author{
Hao Liu$^1$
\and
Yu Hu$^1$\and
Rakiba Rayhana$^1$\and
Ling Bai$^1$\And
Zheng Liu$^1$\\
\affiliations
$^1$The University of British Columbia, Canada\\
\emails
\{hao.liu, yu.hu, rakiba.rayhana, ling.bai, zheng.liu\}@ubc.ca
}
\begin{document}

\maketitle

\begin{abstract}
Bed-related falls remain a major source of injury in hospitals and long-term care facilities, yet many commercial alarms trigger only after a patient has already left the bed. We show that early bed-exit intent can be predicted using only one low-cost load cell mounted under a bed leg. The resulting load signals are first converted into a compact set of complementary images: an RGB line plot that preserves raw waveforms and three texture maps—recurrence plot, Markov transition field, and Gramian angular field—that expose higher-order dynamics. We introduce ViFusionTST, a dual-stream Swin Transformer that processes the line plot and texture maps in parallel and fuses them through cross-attention to learn data-driven modality weights. To provide a realistic benchmark, we collected six months of continuous data from 95 beds in a long-term-care facility. On this real-world dataset ViFusionTST reaches an accuracy of 0.885 and an F1 score of 0.794, surpassing recent 1D and 2D time-series baselines across F1, recall, accuracy, and AUPRC. The results demonstrate that image-based fusion of load-sensor signals for time series classification is a practical and effective solution for real-time, privacy-preserving fall prevention.

\end{abstract}

\section{Introduction}
Falls are a frequent and costly issue among older adults in hospitals and long-term care facilities, with U.S. hospitals reporting 700,000 to 1 million falls annually, and up to 10\% resulting in serious injury or death \cite{lelaurin2019preventing}. Many falls occur when patients try to leave their beds unassisted \cite{hitcho2004characteristics,rose2020toileting}, making reliable bed-exit alarm systems crucial for fall prevention \cite{seow2022effectiveness}.

Conventional alarms typically use pressure mats or infrared beams, which often alert caregivers too late, after the patient has already exited the bed \cite{capezuti2009bed}. Recent research has shifted toward predicting bed-exit intentions by detecting pre-exit behaviors, allowing timely warnings to caregivers \cite{meng2024method}.

Various sensing methods, including wearables \cite{awais2019internet}, cameras \cite{Chen2018,Bu2021,hao2014prediction}, pressure mats \cite{meng2024method,kuwahara2017bed,bai2023deriving}, and vibration sensors \cite{valero2021health}, have been explored for bed-exit prediction, but each has practical deployment challenges. Wearables can irritate skin and need frequent charging, making them unsuitable for continuous monitoring. Cameras pose privacy issues and require significant storage and processing power, while also struggling with lighting changes and occlusion \cite{Bisogni2022TII}. Pressure mats offer detailed pressure information but are costly and can reduce comfort, especially on adjustable beds. Vibration sensors can be easily disrupted by ambient noises. Consequently, most studies focus on lab experiments rather than real-world applications.

We argue that effective bed-exit intention prediction can be achieved without complex sensor systems. Bed-leg-mounted load sensors offer a low-cost, non-intrusive, and privacy-friendly solution by continuously monitoring vertical force to capture movements leading up to a bed exit, potentially providing sufficient data for accurate real-time prediction. 
In this study, we convert the one-dimensional (1D) load sensor signals into a set of complementary two-dimensional (2D) images and utilize powerful vision models for time-series modeling, aiming to predict bed exits accurately in real time.

Traditionally, 1D sequence models, ranging from multi-layer perceptrons (MLPs) to deep convolutional neural networks (CNNs)/recurrent neural networks (RNNs), were developed expressly for time-series signals such as bed loads \cite{Zeng2022DLinear,Bai2018TCN}. In recent years, the field has progressed to Time Series Transformers \cite{zerveas2021transformer} and even large language models (LLMs) \cite{Jin2024}. Meanwhile, Vision Transformers \cite{Dosovitskiy2021,Liu2021} have achieved remarkable success in computer vision, motivating researchers to explore whether the same architectures can model temporal data \cite{chen2024visionts}. This exploration is enabled by a rich set of techniques that turn 1D time series into 2D images: line plots for multivariate signals and, for univariate signals, recurrence plots (RP), Markov transition fields (MTF), and Gramian angular fields (GAF) \cite{Li2023,costa2024fusion}. Compared with raw numeric sequences, these images give both humans and neural networks a more intuitive view of temporal patterns. Vision methods for time series bring two practical benefits: 1) by pretraining on a large number of images, vision models have already learned common patterns such as trends, cycles, and peaks; 2) representing a long sequence as a compact image reduces input length, easing the burden of modeling long-range dependencies \cite{ni2025harnessing}.

Yet most prior work \cite{barra2020deep,Li2023,jin2023classification,karami2024timehr} relies on only one image modality—typically a line plot, RP, or GAF—to represent the entire sequence, which limits the information available to the model. FIRTS \cite{costa2024fusion} combines RP, MTF, and GAF channels, but it omits the raw-waveform line plot, and its simple channel summation cannot fully integrate the complementary cues each modality provides. These gaps motivate the multi-modal fusion strategy we adopt in this study.

We propose ViFusionTST, a dual-stream model that processes (i) the line-plot RGB image and (ii) the three texture maps (RP, MTF, GAF) in two parallel Swin encoders, then fuses them through cross-attention to learn optimal modality weights automatically. Our design preserves the low-cost, privacy-friendly sensing hardware while leveraging state-of-the-art vision models for bed monitoring. To build a realistic benchmark, we collected 6 months of data from 95 beds in a long‑term care facility. On this real-world dataset, ViFusionTST outperforms recent state‑of‑the‑art 1D and 2D time series baselines in F1 score, recall, accuracy, and AUPRC. These experimental results demonstrate that ViFusionTST advances image-based time-series modeling and significantly improves prediction performance on the bed-exit task.

\section{Related Work}
\subsection{Bed Exit Prediction}
Predicting a bed exit before it happens remains difficult in healthcare practice. Most existing systems only raise an alarm after the patient leaves the bed \cite{Lin2022,Chang2021,Inoue2020}, leaving caregivers little time to act.

Early studies have sought to bridge this gap by employing camera-based human-action recognition methods. Chen et al. \cite{Chen2018} used a depth camera and a 3D CNN to detect two pre-exit motions: torso rotation while sitting and rolling out of bed. However, their study provided only qualitative examples of predictive capability without a quantitative analysis of its accuracy.
BED-net \cite{Bu2021} extended this idea with an R(2+1)D backbone and reached 87.7\% accuracy detecting transitional movements, yet performance dropped when the camera angle changed or blankets covered the subject. 

Other researchers have explored pressure mats or vibration sensors. Meng et al. \cite{meng2024method} trained a 1-D CNN on air-spring mattress pressure data and achieved 93.5 \% overall accuracy in a laboratory setting, but did not report recall or precision. Valero et al. \cite{valero2021health} monitored bed vibrations to recognize one pre-exit posture - sitting. The system was not trained or evaluated to detect other exit patterns, limiting its practical scope.

Across different modalities, existing methods either expose sensitive imagery of patients, require expensive hardware, or cover only a limited set of bed-exit behaviors. We present the first bed-exit prediction system that works with a single low-cost load cell mounted under one bed leg. The sensor fits any bed frame, adds minimal hardware cost, and scales easily. Moreover, our model learns the common cues across diverse bed-exit motions directly from data, without hard-coding specific transitions, making it flexible for real-world use.

\subsection{Deep Learning for Time‑Series Prediction}
\subsubsection{1D Time Series}
Time series data are most naturally and commonly represented as 1D sequences, where each timestamp corresponds to a vector measurement in the multivariate case. This representation directly models raw signals, thus it is widely applied in diverse time series tasks.

Transformers, since their introduction in 2017 by Vaswani et al. \cite{Vaswani2017}, have revolutionized a wide range of domains, including time series analysis \cite{Wen2023}. Notable adaptations include TST, which applies the vanilla Transformer encoder to multivariate sequences \cite{zerveas2021transformer}; iTransformer, which treats each variable as a token to emphasize cross-variable interactions \cite{Liu2023iTransformer}. Meanwhile, lightweight non-attention architectures—MLP-based (e.g. DLinear, TSMixer \cite{Zeng2022DLinear,Chen2023TSMixer}), CNN-based (e.g. TCN, TimesNet \cite{Bai2018TCN,Wu2023TimesNet}), and state-space-based (e.g. Mamba \cite{Gu2023Mamba})—discard global attention to reduce latency and memory usage, yet still achieve competitive benchmark results.

Despite these advances, 1D methods struggle to capture complex hidden patterns in noisy signal data. These methods also do not perform well on long context windows and irregularly sampled sequences. This limitation motivates mapping time series to 2D images to exploit pretrained vision models' spatial representation ability.

\subsubsection{Time Series as 2D Images}

2D transformations such as line plot, RP, MTF, and GAF map temporal dynamics into images, turning temporal patterns into spatial textures. These transformations let us exploit the powerful spatial feature extraction of pretrained vision backbones including ResNet \cite{He2015}, ViT \cite{Dosovitskiy2021} and Swin Transformer \cite{Liu2021}. These methods have proven useful by previous studies, especially in classification tasks. FIRTS \cite{costa2024fusion} converts univariate time series into RP, MTF, and GAF then sums these three representations into a single grayscale image and uses GoogLeNet for time series classification. \cite{jin2023classification} use CNNs on line plot images for univariate classification, and ViTST \cite{Li2023} extends this approach by applying line plot images and a Swin Transformer to multivariate and irregularly sampled time series. ViTST outperforms 1D methods on irregular data while remaining competitive on regular sequences.

Although single-modality line plots are simple to generate, they fail to capture the full information of temporal dynamics. The RP, MTF, and GAF modalities provide complementary insights, but naive summation by FIRTS blends their information indiscriminately, which is insufficient to fully fuse and balance different modality contributions. In contrast, ViFusionTST employs dual Swin encoders to extract spatial features from line plot, RP, MTF, and GAF modalities and applies cross-attention fusion to learn optimal modality weights, integrating complementary information for enhanced performance on the time series task.

\section{Methodology}
\subsection{Bed Monitoring Platform Overview}
\begin{figure}
    \centering
    \includegraphics[width=1\linewidth]{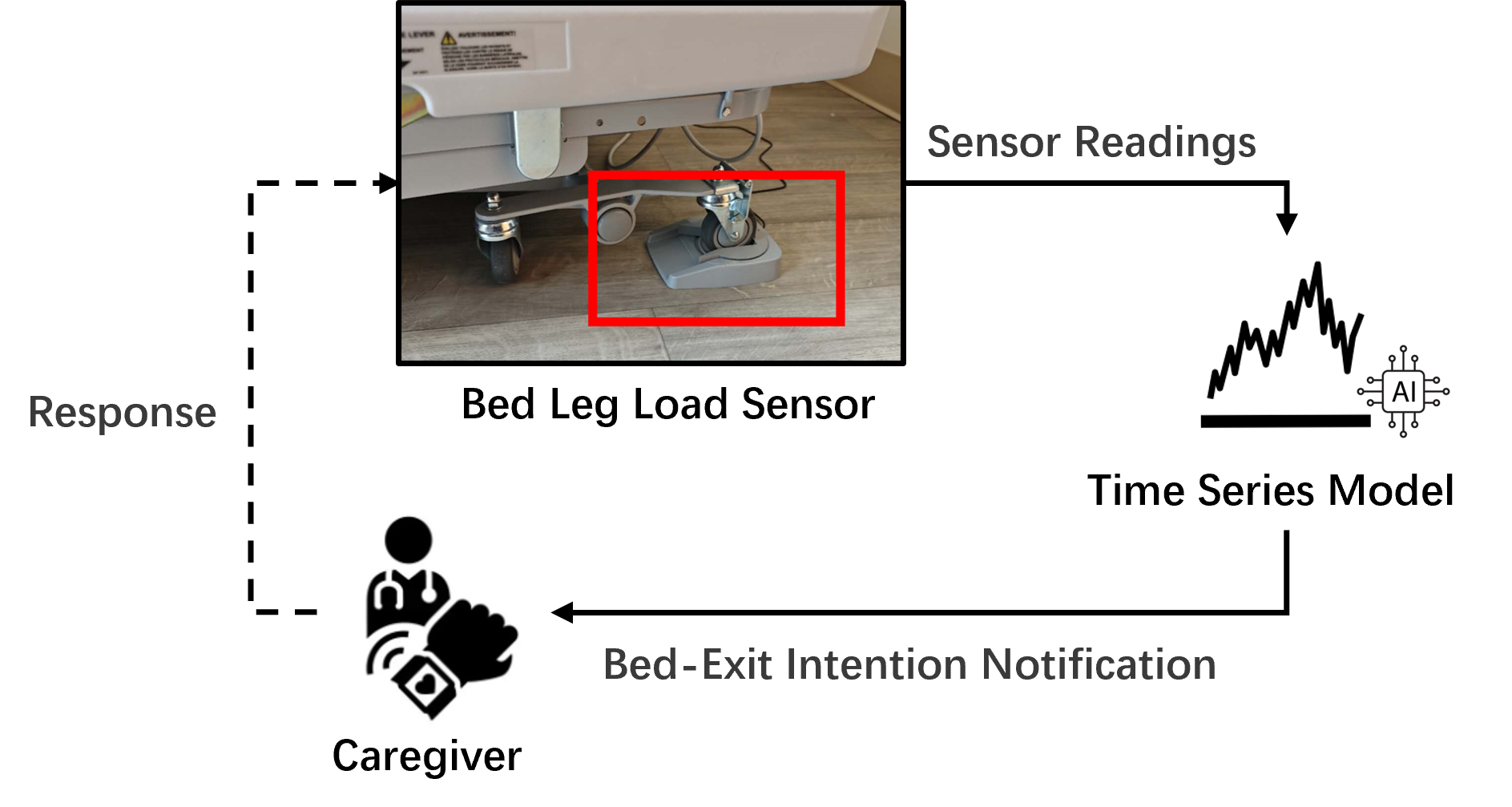}
    \caption{Overview of the strain gauge-based bed monitoring platform.}
    \label{fig:monitoring_platform}
\end{figure}

In this study, we develop a bed monitoring platform that enables unobtrusive, non‐contact monitoring of hospital beds, as illustrated in Fig. \ref{fig:monitoring_platform}. The bed sensor employs a strain gauge–based load cell housed in an enclosure with a built‐in ramp and platform, allowing a bed wheel to roll up and remain securely on top. The sensor is powered by a standard power cord and transmits data wirelessly to our AI platform through an IoT (Internet of Things) network.

The bed sensor records three primary signals: load, vibration, and bed occupancy status. The load signal is directly measured by the load cell. The vibration signal is derived from the load signal using a band-pass filter between 0.5 Hz and 10 Hz. Bed occupancy status is determined by an adaptive thresholding method, which dynamically calculates and distinguishes between empty bed and occupied bed weights, thus accurately indicating bed occupancy.

\begin{figure*}
    \centering
    \includegraphics[width=1\linewidth]{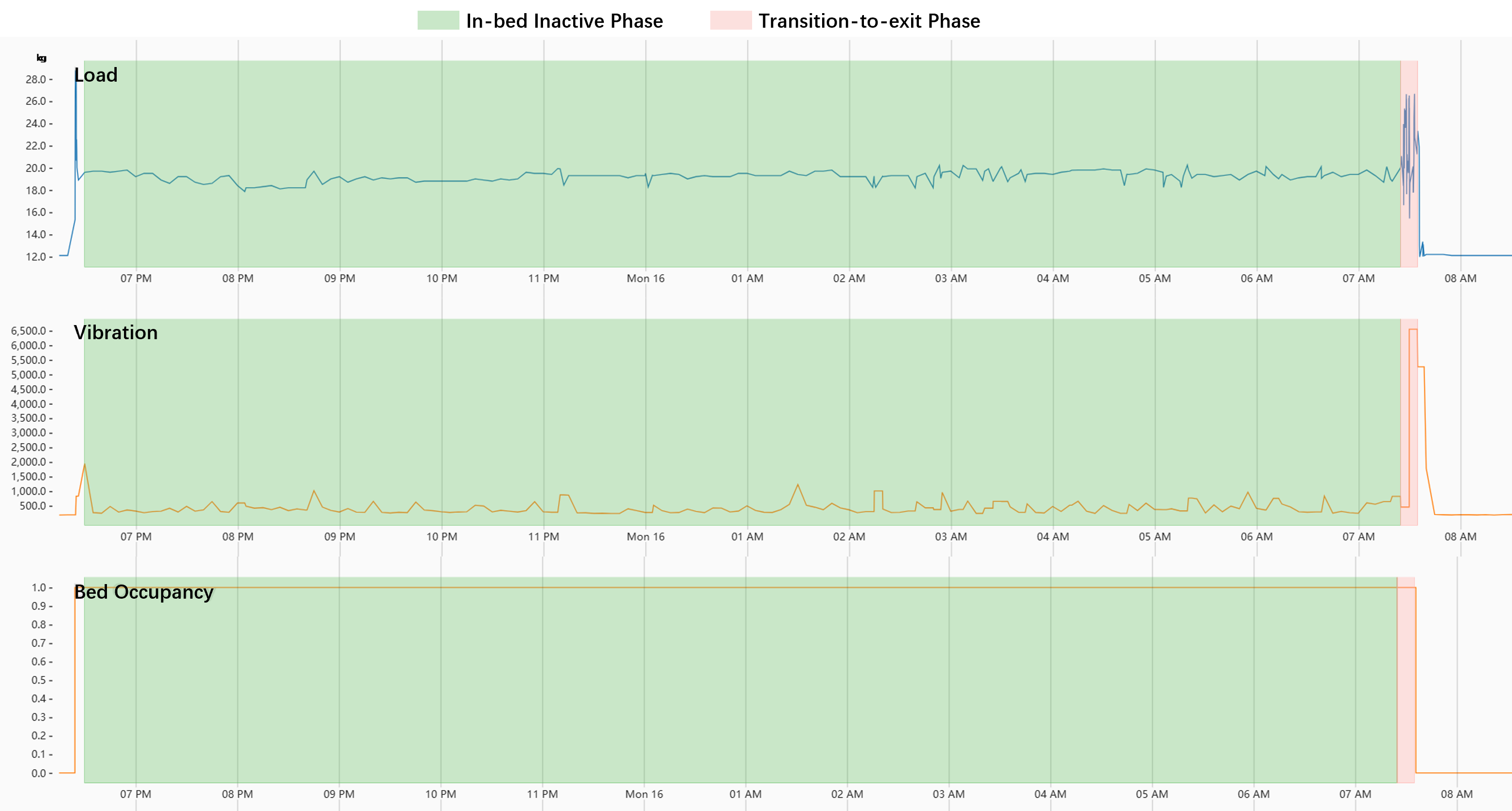}
    \caption{Sensor signals capturing the in-bed period and bed-exit transition.}
    \label{fig:signal_sample}
\end{figure*}

In healthcare practice, patients frequently exhibit a distinct transition phase before leaving their beds, as demonstrated in Fig.~\ref{fig:signal_sample}. During this transition, patients typically move towards the bed edge, causing noticeable fluctuations in load sensor measurements and heightened vibration signals. The transition period generally spans from 1 to 10 minutes prior to an actual bed exit, varying according to individual patient physical abilities. Conversely, non-active periods represent phases of minimal movement or minor repositioning, occupying the majority of the in-bed duration, and preceding the transition phase and subsequent bed exit.

The primary objective of our approach, implemented through the proposed ViFusionTST model, is to accurately distinguish between genuine pre-exit behaviors and routine movements such as repositioning or resting. ViFusionTST analyzes patterns within the load sensor signals to identify transitions indicating a patient's intention to exit the bed. Upon detecting these pre-exit patterns, the system promptly alerts caregivers through connected devices, facilitating timely interventions.

\subsection{Image Representations of Time Series}
To use powerful modern pretrained vision backbones as spatial feature extractors for modeling complex temporal correlations, we transform the 1D time-series signals from the bed monitoring platform into various 2D image representations. Specifically, we employ four distinct transformations: Line Plot, RP, MTF, and GAF. Each representation captures unique aspects of the temporal data, enhancing the model’s ability to identify critical patterns associated with bed-exit behavior.
\subsubsection{Line Plot}
The line plot representation offers a direct, human‐intuitive visualization of multivariate time series, similar to how one inspects temporal patterns by eye. Following the ViTST methodology \cite{Li2023}, we render each time points as a marker and connect markers with straight lines. We visualize each variable separately as distinct subplots and differentiate them by color. In our study, besides the three primary signals—load, vibration, and bed occupancy status—we additionally calculate a derived feature, in-bed duration, as a fourth variable to capture longer-term occupancy behavior. These four variables are arranged into a \(2\times2\) subplot configuration, as illustrated in Fig.~\ref{fig:model}.

\subsubsection{Recurrence Plot (RP)}
RP \cite{eckmann1995recurrence} encodes the times at which a dynamical system revisits a similar state. Given a univariate time series \(\{x_t\}_{t=1}^N\), we first use the 1D values directly to reconstruct a phase‐space trajectory. The recurrence matrix \(\mathbf{R}\in\{0,1\}^{N\times N}\) is defined as
\begin{equation}
    R_{i,j} \;=\; 
    \Theta\!\bigl(\varepsilon - \lVert x_i - x_j\rVert\bigr)
    \,,\label{eq:rp}
\end{equation}
where \(\Theta(\cdot)\) is the Heaviside step function and \(\varepsilon\) is a distance threshold.  A black pixel at \((i,j)\) indicates that the states at times \(i\) and \(j\) are within \(\varepsilon\) of each other.
\subsubsection{Markov Transition Field (MTF)}
MTF \cite{wang2015encoding,wang2015imaging} captures the transition probabilities of quantized states over time, projected back onto the temporal axis.  We first quantize each \(x_t\) into one of \(Q\) discrete bins by
\begin{equation}
b_t = \mathrm{bin}\bigl(x_t\bigr)\in\{1,2,\dots,Q\}.
\end{equation}
We estimate the \(Q\times Q\) state‐transition matrix \(\mathbf{P}\), where
\begin{equation}
    P_{k,\ell} \;=\; \Pr\bigl(b_{t+1}=\ell \mid b_t=k\bigr)
    \quad (k,\ell=1,\dots,Q).
\end{equation}
The MTF is then the \(N\times N\) matrix \(\mathbf{M}\) with entries
\begin{equation}
    M_{i,j} \;=\; P_{\,b_i,\,b_j}\,,
    \label{eq:mtf}
\end{equation}
so that each element encodes the transition likelihood between the quantized values at times \(i\) and \(j\).
\subsubsection{Gramian Angular Field (GAF)}
GAF \cite{wang2015encoding,wang2015imaging} represents time-series values in polar coordinates and captures pairwise angular relationships.  First, we rescale the series \(x_t\) to \([-1,1]\), obtaining \(\tilde x_t\).
Then encode each \(\tilde x_t\) as an angle
\begin{equation}
    \phi_t = \arccos(\tilde x_t)\in[0,\pi]\,,\quad t=1,\dots,N.
\end{equation}

The Gramian Angular Summation Field (GASF) is defined by
\begin{equation}
    G_{i,j}^{\mathrm{(sum)}} = \cos\bigl(\phi_i + \phi_j\bigr)\,.
    \label{eq:gaf_sum}
\end{equation}
The resulting \(N\times N\) matrix encodes global temporal dependencies via angular correlations between every pair of time points.

Since the RP, MTF, and GAF methods inherently apply to univariate time series, we select the most informative feature, load, for these transformations. The resulting three representations—RP, MTF, and GAF—are combined as distinct image channels to form a unified 3-channel input to our deep learning model, as shown in Fig.~\ref{fig:model}.

\subsection{ViFusionTST: Deep Fusion for Bed Exit Prediction}
\begin{figure}
    \centering
    \includegraphics[width=1\linewidth]{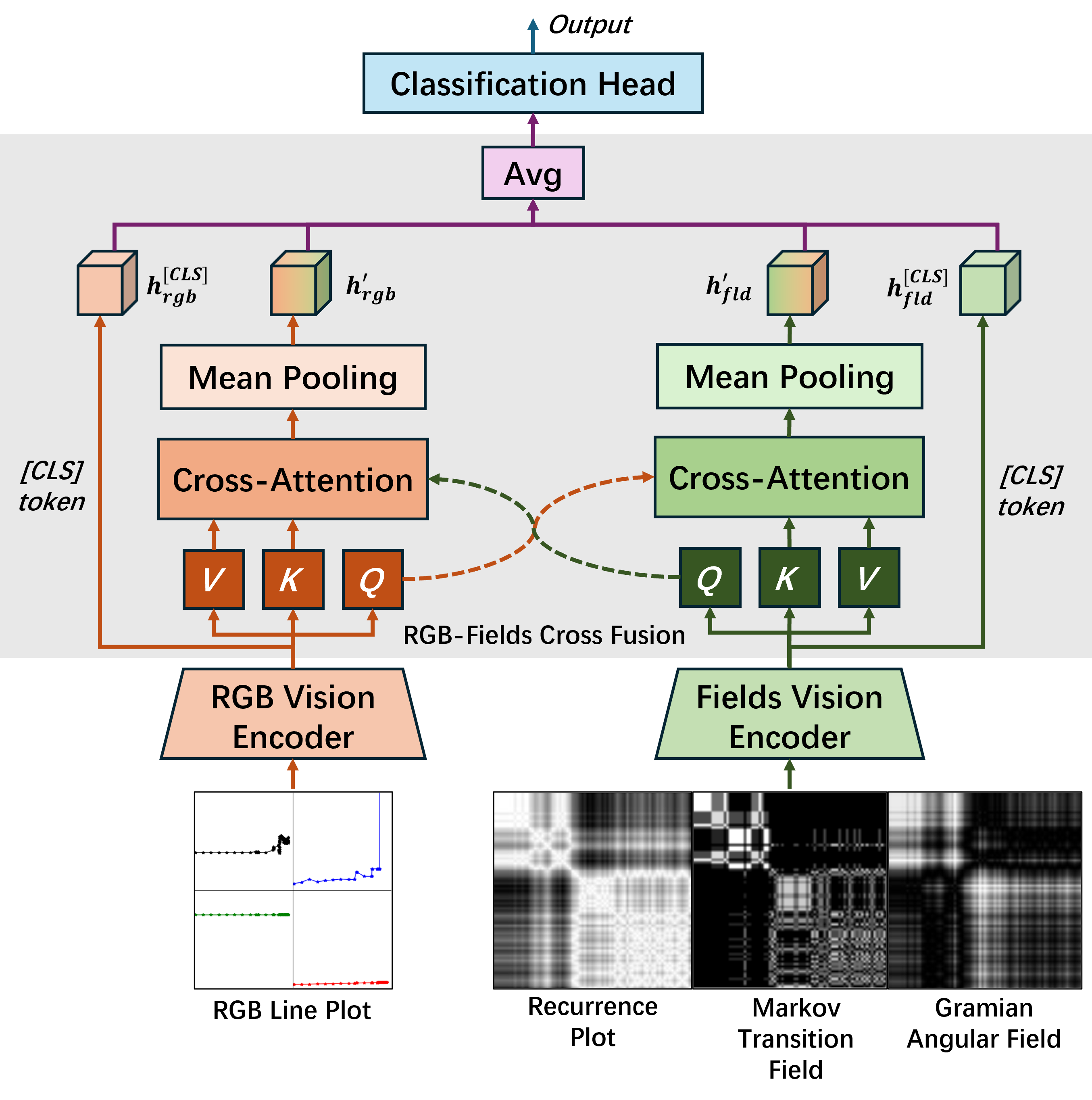}
    \caption{ViFusionTST model architecture.}
    \label{fig:model}
\end{figure}

Given the image representations derived from the time-series signals, we propose ViFusionTST, a deep fusion model that consists of two parallel Swin Transformer vision encoders, each processing distinct image representation modalities, and a deep cross-attention fusion module to effectively integrate their complementary information. Specifically, one vision encoder processes the line plot composed of RGB channels, while the other processes a three-channel image created from the RP, MTF, and GAF. The overall model architecture of ViFusionTST is illustrated in Fig.~\ref{fig:model}.

The motivation for this design comes from the observation that line plots and field-based representations emphasize different temporal characteristics. Line plots inherently preserve the raw waveform details—such as peaks, slopes, and local fluctuations—allowing vision models to interpret these patterns akin to spatial textures and edges in conventional images. In contrast, field-based encodings (RP, MTF, and GAF) reveal higher-order temporal dynamics, including state recurrences, transition probabilities between quantized states, and global angular correlations, respectively. By assigning each modality group to its own vision encoder, we let each network specialize on the unique statistical patterns in its input. The encoder handling RGB line plots specializes in capturing fine-grained, local temporal patterns, whereas the encoder processing RP, MTF, and GAF representations is optimized for detecting global and structural dependencies within the time series. Subsequently, a cross-attention fusion module dynamically re-weights and integrates the extracted features from both streams, producing joint representations.

Moreover, our approach utilizes the Swin-Tiny Transformer encoder rather than the Swin-Base backbone used in ViTST. We empirically observed that Swin-Tiny is better suited for simpler image representations derived from time series data, significantly mitigating overfitting while reducing computational overhead. This choice achieves a beneficial balance between model complexity and performance, making ViFusionTST efficient, accurate, and practical for real-world bed-exit prediction scenarios.

% simple but effective cross fusion method

\section{Experiments and Results}
In this section, we describe our experimental setup and evaluate our model against state-of-the-art methods. We also conduct ablation and performance analysis to validate our approach.
\subsection{Experimental Setup}
\subsubsection{Data Collection \& Preparation}
For this study, we collected sensor readings from our bed sensor platform deployed across 95 beds at a long-term care facility in British Columbia, Canada. Data was recorded over a period spanning December 21, 2023, to May 30, 2024. To ensure a diverse representation of patient behaviors, mobility levels, and sensor configurations, we manually reviewed and selected data from 57 beds, totaling 579 days of sensor readings for labeling.

The manual labeling process focused on accurately defining transition periods and non-active periods, as illustrated in Fig.~\ref{fig:signal_sample}. A total of 1\,211 transition periods were annotated, along with corresponding non-active periods. These samples were then split 60\%/20\%/20\% for training, validation, and testing, respectively.

\subsubsection{Implementation Details}

The proposed ViFusionTST is implemented in Pytorch and trained using the AdamW optimizer with a batch size of 256 and a learning rate of \(2 \times 10^{-5}\). Input image representations of time series, including line plots, RP, MTF, and GAF, are all resized to 224×224 to match the Swin-tiny backbone pretrained on ImageNet. A 3-hour look-back window is employed for all time series inputs. We apply early stopping with a patience of 10 evaluation steps, and the model checkpoint with the highest validation accuracy is then used for the final test evaluation. All experiments are conducted on a machine equipped with dual NVIDIA RTX 2080 Ti GPUs.

% \subsubsection{Validation Metrics}
% To evaluate our model’s predictive performance on bed-exit intention, we use the following standard classification metrics. Let \(\mathrm{TP}\), \(\mathrm{TN}\), \(\mathrm{FP}\), and \(\mathrm{FN}\) denote the numbers of true positives, true negatives, false positives, and false negatives, respectively.

% \begin{align}
% \text{Accuracy} &= \frac{\mathrm{TP} + \mathrm{TN}}{\mathrm{TP} + \mathrm{TN} + \mathrm{FP} + \mathrm{FN}}\,, \\[6pt]
% \text{Precision} &= \frac{\mathrm{TP}}{\mathrm{TP} + \mathrm{FP}}\,, \\[6pt]
% \text{Recall} &= \frac{\mathrm{TP}}{\mathrm{TP} + \mathrm{FN}}\,, \\[6pt]
% \text{F1 Score} &= 2 \times \frac{\text{Precision} \times \text{Recall}}{\text{Precision} + \text{Recall}}\,.
% \end{align}

% \noindent
% We also report the \emph{Area Under the Precision–Recall Curve (AUPRC)}, which in practice is approximated by a Riemann sum over discrete operating points:
% \begin{equation}
%     \mathrm{AUPRC} \approx \sum_{i=1}^{N} \bigl(R_i - R_{i-1}\bigr)\,P_i\,,
% \end{equation}
% where \((P_i, R_i)\) are the precision–recall pairs sorted by increasing recall, for \(i=1,\dots,N\).

\subsection{Comparative Study}
To evaluate the effectiveness of our ViFusionTST model for bed-exit prediction, we benchmarked it against 10 state-of-the-art methods: eight 1D time-series models—iTransformer \cite{Liu2023iTransformer}, TST \cite{zerveas2021transformer}, DLinear \cite{Zeng2022DLinear}, LightTS \cite{zhang2022morefastmultivariatetime}, TSMixer \cite{Chen2023TSMixer}, Temporal Convolutional Networks (TCN) \cite{Bai2018TCN}, TimesNet \cite{Wu2023TimesNet}, and Mamba \cite{Gu2023Mamba}; two 2D time-series models—ViTST \cite{Li2023} and FIRTS \cite{costa2024fusion}. 

The comparative results are presented in Table.~\ref{tab:model_comparison}. The best results are in {\color{red} \textbf{red}}, and the second-best results are {\color{blue} \underline{underlined}}. Our ViFusionTST model outperforms previous baseline models on 4 out of 5 metrics. Specifically, our method achieves an F1 score of 79.4\%, which is 6.0 pp higher than the second-best ViTST (73.4\%), corresponding to an 8.2\% relative improvement. Similarly, recall increases by 14.8 pp (21.8\%), accuracy by 2.1 pp (2.4\%), and AUPRC by 3.6 pp (4.3\%). Meanwhile, our fusion-based method maintains high precision, preventing excessive false alerts and thus avoiding alarm fatigue among caregivers.

\begin{table*}
    \centering
    \begin{tabular}{c|l|>{\centering\arraybackslash}m{1.6cm}|>{\centering\arraybackslash}m{1.4cm}|>{\centering\arraybackslash}m{1.7cm}|>{\centering\arraybackslash}m{1.7cm}|>{\centering\arraybackslash}m{1.5cm}}
        \toprule
        \textbf{Dim.} & \textbf{Model} & \textbf{F1 Score $\uparrow$} & \textbf{Recall $\uparrow$} & \textbf{Precision $\uparrow$} & \textbf{Accuracy $\uparrow$} & \textbf{AUPRC $\uparrow$} \\
        \midrule
        \multirow{10}{*}{1D} & DLinear       & 0.584 & 0.519 & 0.667 & 0.802 & 0.686 \\
                            & LightTS       & 0.718 & 0.678 & 0.764 & 0.858 & 0.801 \\
                            & TSMixer       & 0.733 & 0.701 & 0.769 & 0.863 & 0.825 \\
                            & TCN           & 0.688 & \underline{\textcolor{blue}{0.738}} & 0.644 & 0.821 & 0.721 \\
                            & TimesNet      & 0.690 & 0.631 & 0.759 & 0.848 & 0.750 \\
                            & Mamba         & 0.701 & 0.631 & \underline{\textcolor{blue}{0.787}} & 0.855 & 0.787 \\
                            & iTransformer  & 0.730 & 0.707 & 0.754 & 0.860 & 0.807 \\
                            & TST           & 0.716 & 0.672 & 0.766 & 0.857 & 0.809 \\
        \midrule
        \multirow{3}{*}{2D}  & FIRTS         & 0.659 & 0.640 & 0.680 & 0.823 & 0.783 \\
                            & ViTST         & \underline{\textcolor{blue}{0.734}} & 0.679 & \textbf{\textcolor{red}{0.797}} & \underline{\textcolor{blue}{0.868}} & \underline{\textcolor{blue}{0.832}} \\
                            & \textbf{ViFusionTST (Ours)} & \textbf{\textcolor{red}{0.794}} & \textbf{\textcolor{red}{0.827}} & 0.763 & \textbf{\textcolor{red}{0.885}} & \textbf{\textcolor{red}{0.868}} \\
        \bottomrule
    \end{tabular}
    \caption{Performance comparison of different models for bed-exit prediction.}
    \label{tab:model_comparison}
\end{table*}

This consistent gain results from our cross-attention fusion module, which dynamically re-weights features from both line plots and RP/MTF/GAF representations, leading to more discriminative embeddings during short transition periods. Compared to FIRTS which relies solely on RP/MTF/GAF, and ViTST which processes only line plots, our multi-modal fusion approach substantially increases recall, thereby improving overall predictive performance. Moreover, both ViTST and our ViFusionTST—which incorporate line-plot representations as input—outperform the 1D baselines. We attribute this to the vision-based architectures’ better ability to capture long-term dependencies from extended irregular context windows in line plots—patterns that 1D models struggle to learn from raw time-series data.

\subsection{Ablation Study}
% Compare 1 encoder vs. 2 encoder fusion; rgb, rf, gaf, mtf ablation;
To quantify the individual contributions of our fusion strategy and the choice of input modalities, we carried out two ablation experiments. First, we compared our cross-attention fusion against a simple feature-concatenation baseline, and then evaluated models trained on different input representations.

\subsubsection{Effect of Fusion Method}
To isolate the impact of our cross‐attention fusion, we compare against a feature‐concatenation baseline that uses the identical backbone and training protocol as ViFusionTST. As shown in Table \ref{tab:fusion_methods}, replacing feature concatenation with our cross-attention fusion module yields large gains across all metrics. In particular, F1 score increases from 0.658 to 0.794 (+0.136, +20.7\%), recall from 0.582 to 0.827 (+0.245, +42.1\%), and AUPRC from 0.761 to 0.868 (+0.107, +14.1\%). This confirms that our cross-attention module more effectively integrates complementary features than naive concatenation.

\begin{table*}[h!]
    \centering
    \begin{tabular}{l|ccccc}
        \toprule
        \textbf{Input Modality} & \textbf{F1 Score $\uparrow$} & \textbf{Recall $\uparrow$} & \textbf{Precision $\uparrow$} & \textbf{Accuracy $\uparrow$} & \textbf{AUPRC $\uparrow$} \\
        \midrule
        Line Plot                       & 0.771 & 0.824 & 0.725 & 0.869 & 0.835 \\
        RP/MTF/GAF                      & 0.659 & 0.640 & 0.680 & 0.823 & 0.732 \\
        \textbf{Line Plot + RP/MTF/GAF (Ours)}  & \textbf{0.794} & \textbf{0.827} & \textbf{0.763} & \textbf{0.885} & \textbf{0.868} \\
        \bottomrule
    \end{tabular}
    \caption{Impact of different input Modalities on prediction performance.}
    \label{tab:input_modality}
\end{table*}

\begin{table*}[ht!]
    \centering
    \begin{tabular}{l|cc|cc|cc|cc|cc}
        \toprule
        \multirow{2}{*}{\textbf{Fusion Method}} 
            & \multicolumn{2}{c|}{\textbf{F1 Score $\uparrow$}} 
            & \multicolumn{2}{c|}{\textbf{Recall $\uparrow$}} 
            & \multicolumn{2}{c|}{\textbf{Precision $\uparrow$}} 
            & \multicolumn{2}{c|}{\textbf{Accuracy $\uparrow$}} 
            & \multicolumn{2}{c}{\textbf{AUPRC $\uparrow$}} \\
        \cmidrule{2-11}
            & Value & Diff & Value & Diff & Value & Diff & Value & Diff & Value & Diff \\
        \midrule
        Early Concatenation         & 0.658 & -17.1\% & 0.582 & -29.6\% & 0.756 & -0.9\%  & 0.838 & -5.3\% & 0.761 & -12.3\% \\
        Gated Fusion                & 0.722 & -9.1\%  & 0.672 & -18.7\% & 0.780 & +2.2\%  & 0.861 & -2.7\% & 0.814 & -6.2\% \\
        Mid Concatenation           & 0.764 & -3.8\%  & 0.720 & -12.9\% & 0.814 & +6.7\%  & 0.881 & -0.5\% & 0.847 & -2.4\% \\
        \textbf{Cross Fusion (Ours)} & \textbf{0.794} & ref. & \textbf{0.827} & ref. & \textbf{0.763} & ref. & \textbf{0.885} & ref. & \textbf{0.868} & ref. \\
        \bottomrule
    \end{tabular}
    \caption{Evaluation of fusion methods. “Diff” columns indicate relative difference compared to Cross Fusion (Ours).}
    \label{tab:fusion_methods}
\end{table*}

\begin{table*}[ht]
  \centering
  \begin{tabular}{l|ccccc}
    \toprule
    \textbf{Vision Encoder} & \textbf{F1 Score $\uparrow$} & \textbf{Recall $\uparrow$} & \textbf{Precision $\uparrow$} & \textbf{Accuracy $\uparrow$} & \textbf{AUPRC $\uparrow$} \\
    \midrule
    Resnet-50 \cite{He2015}              & 0.727 & 0.663 & 0.804 & 0.866 & 0.817 \\
    \midrule
    % MambaVision 
    MambaVision-Base \cite{hatamizadeh2025mambavision}      & 0.731 & 0.665 & 0.812 & 0.869 & 0.813 \\
    MambaVision-Small \cite{hatamizadeh2025mambavision}     & 0.729 & 0.655 & 0.824 & 0.870 & 0.821 \\
    MambaVision-Tiny \cite{hatamizadeh2025mambavision}      & 0.759 & 0.702 & \textbf{0.825} & 0.880 & 0.830 \\
    \midrule
    % SwinV2 
    SwinV2-Base \cite{liu2022swin}           & 0.701 & 0.656 & 0.753 & 0.850 & 0.785 \\
    SwinV2-Small \cite{liu2022swin}          & 0.734 & 0.815 & 0.668 & 0.842 & 0.782 \\
    \midrule
    % ViT 
    ViT-Large \cite{Dosovitskiy2021}             & 0.721 & 0.649 & 0.811 & 0.865 & 0.783 \\
    ViT-Base \cite{Dosovitskiy2021}              & 0.731 & 0.690 & 0.778 & 0.864 & 0.822 \\
    \midrule
    % Swin 
    Swin-Base \cite{Liu2021}             & 0.766 & 0.797 & 0.738 & 0.870 & 0.817 \\
    Swin-Small \cite{Liu2021}            & 0.775 & 0.783 & 0.767 & 0.878 & 0.822 \\
    \textbf{Swin-Tiny} \cite{Liu2021}      & \textbf{0.794} & \textbf{0.827} & 0.763 & \textbf{0.885} & \textbf{0.868} \\
    \bottomrule
  \end{tabular}
  \caption{Performance Comparison of Vision Encoders}
  \label{tab:vision-encoder-performance}
\end{table*}

\subsubsection{Effect of Input Modalities}
Table \ref{tab:input_modality} reports results for three input configurations: line plots only, RP/MTF/GAF only, and their fusion. The line-plot–only variant achieves an F1 of 0.771 and AUPRC of 0.835, outperforming the RP/MTF/GAF–only variant (F1 = 0.659, AUPRC = 0.732), which indicates that raw curves carry stronger temporal cues. When both modalities are fused, the model reaches F1 = 0.794 (+2.3 pp over line plot, +13.5 pp over RP/MTF/GAF). These results demonstrate that line plots and RP/MTF/GAF representations provide complementary information and that their joint modeling yields the best predictive performance.

\subsubsection{Effect of Vision Encoders}

To quantify the impact of different pretrained vision backbones on bed-exit prediction, we replace our Swin-Tiny encoders with a variety of alternatives, while keeping all other components fixed. Table \ref{tab:vision-encoder-performance} reports the performance for each encoder. The Swin-Tiny encoder achieves the best overall performance, surpassing all other encoders, including its larger variants. Among non-Swin architectures, the MambaVision delivers the next best results, with MambaVision-Tiny achieving F1 = 0.759 and AUPRC = 0.830. This suggests that replacing self-attention layers with Mamba blocks can also effectively capture temporal image features while reducing computational cost.

Notably, within each model family, the smaller variant consistently outperforms its larger counterpart (e.g., ViT-Base vs. ViT-Large). We attribute this to the relatively low complexity of temporal image representations compared to natural images. Larger models tend to overfit on these simpler patterns, leading to degraded generalization. Overall, these findings emphasize the importance of selecting appropriately sized encoders for time-series image inputs, since larger models are not necessarily better.

\subsubsection{Prediction Visualization}

\begin{figure}
    \centering
    \includegraphics[width=1\linewidth]{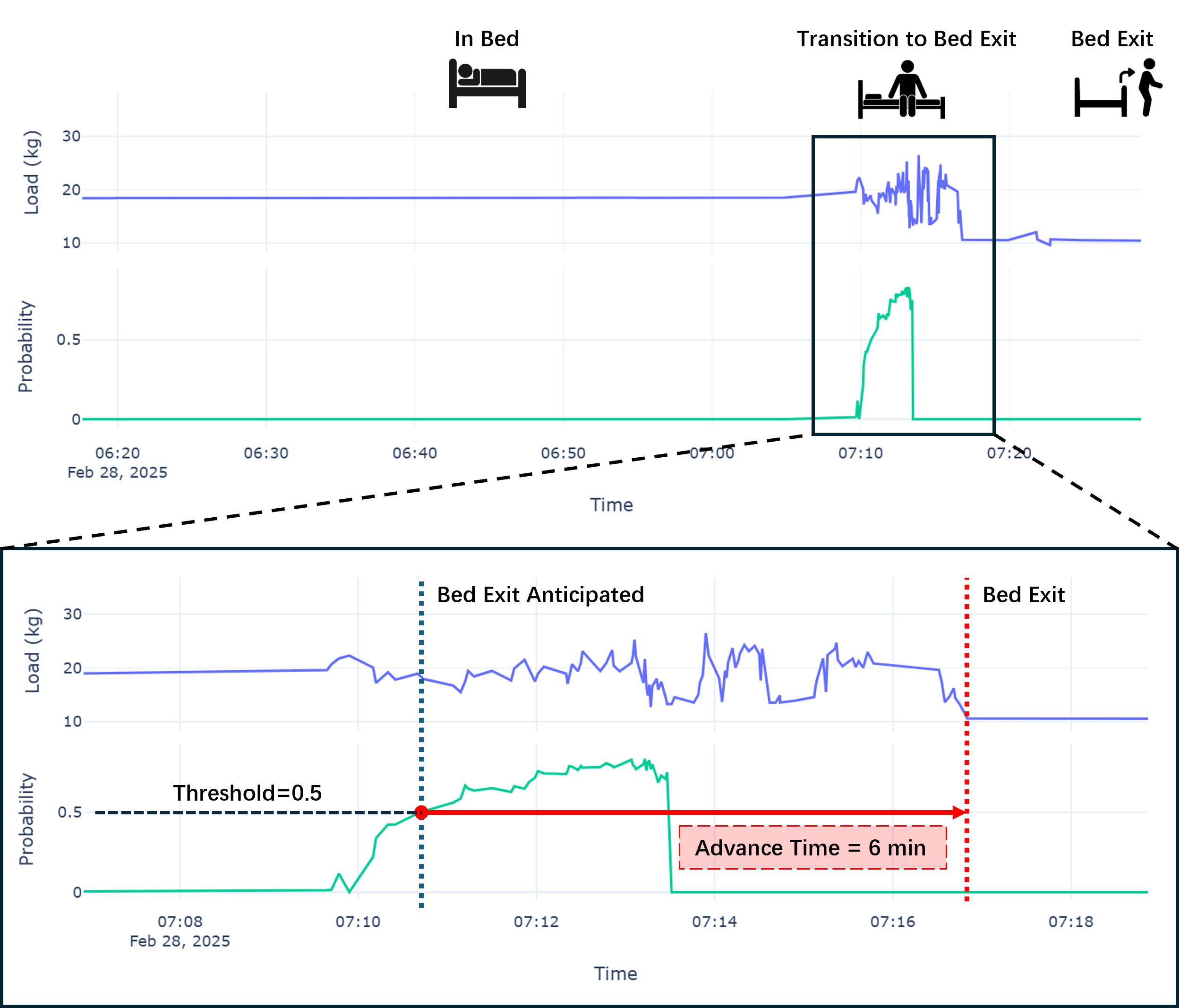}
    \caption{ViFusionTST prediction probability over time for a bed-exit event.}
    \label{fig:pred_vis}
\end{figure}

Figure~\ref{fig:pred_vis} presents a representative bed-exit event, where the load signal (blue) and the predicted bed-exit probability (green) are plotted over time. During the initial stable-in-bed phase, the patient remains motionless. Consequently, the load is flat and the model outputs a probability of~0, suppressing unnecessary alarms. As the patient transitions toward bed exit, subtle posture shifts introduce high-frequency load oscillations. These precursor movements cause the probability to rise steadily. The ward's nurse call system employs a fixed alarm threshold of~0.5 (dashed line). The model first exceeds this threshold six minutes before the true exit time. This advance notice is clinically valuable: caring staff receive a single precise alert rather than a series of premature or fluctuating warnings. These observations demonstrate that the ViFusionTST model can identify preparatory movements that simple load-drop heuristics cannot detect, while providing interpretable and actionable predictive results suitable for deployment in the real world.

\section{Conclusion}
This study shows that a simple load-sensor installation, combined with state-of-the-art vision backbones, is enough to achieve high accuracy, real-time bed-exit prediction in a long-term care setting. We introduced ViFusionTST, a dual-stream Swin Transformer that separately encodes RGB line-plot images and three texture maps (RP, MTF, GAF), then unifies them with cross-attention to learn data-driven modality weights. Evaluated on six months of data from 95 beds, ViFusionTST achieved an accuracy of 0.885 and an F1 score of 0.794, outperforming recent 1D and 2D time-series baselines across all major metrics, establishing a new benchmark for load-sensor–based fall prevention.

Our work contributes (i) a scalable, privacy-preserving sensing pipeline that integrates seamlessly with existing nurse-call systems, (ii) a large real-world benchmark for bed load signal modeling, and (iii) a deep fusion architecture that can be adapted to other regular or irregular time series. Collectively, these advances move image-based time-series modeling closer to real-world deployment and offer a practical path toward reducing fall-related injuries in hospitals and long-term care facilities.

\section*{Acknowledgments}
This work was supported by the Natural Sciences and Engineering Research Council of Canada (NSERC) under grant ALLRP 571106-21, and conducted in collaboration with Tochtech Technologies Ltd., who provided the bed load sensor modules and online data platform for data collection.

\bibliographystyle{named}
\bibliography{ijcai25}

\end{document}